\documentclass{article}

\usepackage[T1]{fontenc}
\usepackage{microtype}
\usepackage{graphicx}
\usepackage{booktabs}
\usepackage{amsmath,amssymb}
\usepackage{hyperref}

\usepackage[preprint]{icml2026}
\icmltitlerunning{Look Ahead: Selective Search for Robotic Control}

\begin{document}

\twocolumn[
\icmltitle{Look Ahead: Search-Driven Reasoning in Embodied AI}

\icmlsetsymbol{equal}{*}
\begin{icmlauthorlist}
\icmlauthor{Ahmed Zeer}{ytu}
\icmlauthor{Osama Shbib}{equal,ytu}
\icmlauthor{Muzaffer Kaan Yuce}{equal,ytu}
\icmlauthor{Himmet Toprak Kesgin}{equal,ytu}
\icmlauthor{Mehmet Fatih Amasyali}{equal,ytu}
\end{icmlauthorlist}

\icmlaffiliation{ytu}{Department of Computer Engineering, Yildiz Technical University, Istanbul, T\"urkiye}
\icmlcorrespondingauthor{Ahmed Zeer}{ahmed.zeer@std.yildiz.edu.tr}
\icmlkeywords{Robot learning, vision-language-action models, action chunking, beam search, test-time compute}
\vskip 0.1in
{\centering\small
\href{https://azeer.co/research/lookahead/}{Project website}\quad
\href{https://github.com/AhmedZeer/lerobot-tree/tree/main/examples/tree_search/pusht}{Code}\par}
\vskip 0.3in
]

\printAffiliationsAndNotice{\icmlEqualContribution}
\hypersetup{
  pdftitle={Look Ahead: Search-Driven Reasoning in Embodied AI},
  pdfauthor={Ahmed Zeer, Osama Shbib, Muzaffer Kaan Yuce, Himmet Toprak Kesgin, Mehmet Fatih Amasyali},
  pdfsubject={Selective test-time search for robotic action selection},
  pdfkeywords={robot learning, VLA, action chunking, beam search, PushT}
}

\begin{abstract}
Search allows a robotic policy to evaluate candidate futures before committing to an action. Motivated by test-time planning for vision--language--action (VLA) models, we study selective search over action chunks in PushT using a frozen Action Chunking with Transformers (ACT) policy. Beam search generates perturbed alternatives around the policy prediction and scores them through rollouts from cloned simulator states. Across 50 evaluation episodes, beam search raises task success from 0\% to 18\% and increases average cumulative reward from 39.44 to 103.2. We introduce \emph{One Step Ahead} (OSA), a gate that simulates the nominal policy chunk and invokes full search only when its predicted coverage drop exceeds a threshold. In the reported ablations, OSA preserves or improves success while reducing runtime by approximately $3\times$ relative to full beam search. Ablations over tree depth, branching factor, beam width, and replanning frequency show that larger search budgets do not consistently improve performance. The results support selective test-time search for action-chunking policies; evaluation on language-conditioned VLAs and physical robots remains future work.
\end{abstract}

\section{Introduction}
Vision--language--action (VLA) models map visual observations and task instructions to robot actions, offering a route toward generalist control \cite{pi0,smolvla,octo}. Predicting plausible actions, however, does not ensure that their consequences will complete a task. In contact-rich manipulation, a locally reasonable movement can displace an object away from its goal, and errors can accumulate over successive action chunks. Test-time search addresses this problem by evaluating candidate futures before the controller commits to a trajectory \cite{vlareasoner,vlaps}.

Policy-guided search combines a learned action prior with a transition model and a scoring function. The policy restricts exploration to plausible actions, the model predicts their effects, and the score identifies promising futures. This can improve action selection without updating policy parameters, but repeated policy queries and simulated transitions increase inference cost. The practical question is therefore when additional search changes the decision enough to justify that cost.

We investigate this question in PushT, a simulated planar manipulation task in which an agent pushes a T-shaped object toward a target pose \cite{pusht}. We augment a pretrained ACT policy \cite{act} with beam search over locally perturbed action chunks. ACT is a vision-based action-chunking policy rather than a language-conditioned VLA. This isolates action selection from language grounding while retaining the proposal-and-rollout interface used in VLA planning.

\begin{figure*}[t]
 \centering
 \includegraphics[width=0.94\textwidth]{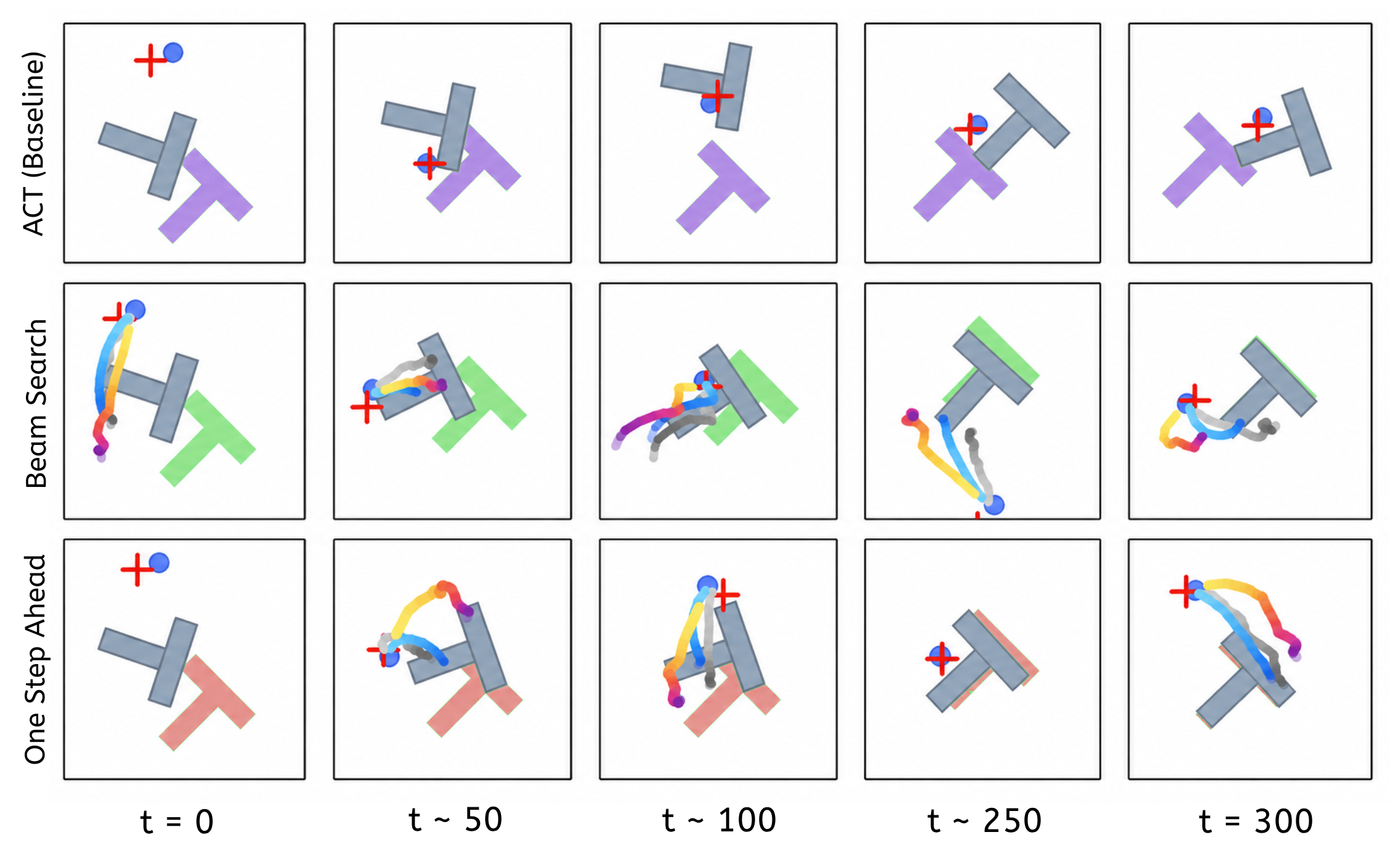}
 \caption{A representative PushT episode under ACT, beam search, and OSA, starting from the same initial configuration. Columns show approximate environment steps. The blue marker is the agent, the gray T is the movable object, and the colored T marks the target. Blue paths show nominal predictions, gray paths show alternatives, and warm-colored paths show selected chunks. The baseline fails to align the object within 300 steps; the search-based controllers reach the target.}
 \label{fig:pusht-rollout-comparison}
\end{figure*}

Full search often returns the same nominal action chunk that the policy would have executed directly. To reduce this overhead, we introduce \emph{One Step Ahead} (OSA). OSA first simulates the nominal chunk and invokes beam search only if its final predicted coverage is lower than the current coverage by more than a fixed threshold. This makes search conditional on a concrete signal of predicted regression. Figure~\ref{fig:pusht-rollout-comparison} illustrates the resulting action selection.

Our contributions are:
\begin{enumerate}
    \item A beam-search controller for PushT that proposes perturbed ACT action chunks, evaluates their consequences using simulator rollouts, and selects a chunk without retraining the policy.
    \item One Step Ahead, a coverage-based gate that selectively invokes search when the nominal policy prediction is expected to reduce task progress.
    \item An evaluation over 50 episodes per configuration, including tree-structure, replanning-frequency, and OSA ablations, with success, reward, alternative-selection, and runtime measurements.
\end{enumerate}

\section{Related Work}
\paragraph{Action-chunking policies and VLAs.}
ACT predicts short sequences of actions from visual observations and robot state using imitation learning \cite{act}. Diffusion Policy represents visuomotor behavior through action diffusion and uses receding-horizon execution \cite{pusht}. Generalist policies extend these ideas to broader task specifications: Octo supports language- and goal-image-conditioned control, $\pi_0$ combines a pretrained vision--language backbone with a flow-based action model, and SmolVLA emphasizes efficient language-conditioned action generation \cite{octo,pi0,smolvla}. Our method operates on the action-chunk interface and keeps the policy frozen. Its evaluation uses ACT, so transfer to language-conditioned policies is an architectural possibility rather than an established result.

\paragraph{Search for VLA policies.}
VLA-Reasoner uses online Monte Carlo tree search with a learned world model and value estimates to evaluate action-conditioned futures \cite{vlareasoner}. VLAPS integrates model-based search with pretrained VLA action priors and evaluates language-specified manipulation in LIBERO \cite{vlaps,libero}. Its experiments assume access to an accurate simulator. These approaches establish the relevance of test-time planning for VLA control. We study a simpler local beam-search procedure in PushT, using continuous coverage feedback and a gate based on the nominal chunk's predicted effect. Our experiments focus on the cost of tree expansion and the value of selective search, rather than a direct performance comparison with these methods on different benchmarks.

\section{Method}
\subsection{Policy and Action Representation}
At control step $t$, the policy receives an observation $o_t=(I_t,s_t)$, where $I_t$ is an RGB image and $s_t\in\mathbb{R}^2$ is the agent position. The frozen policy predicts a nominal chunk
\begin{equation}
 A_t^{(0)}=\pi_{\theta}(o_t)
 = (a_t,\ldots,a_{t+L-1})\in\mathbb{R}^{L\times 2},
 \label{eq:act_policy}
\end{equation}
where each action is a 2D target position in the workspace. We distinguish the candidate rollout length $L$ from the execution horizon $H\leq L$: search evaluates candidate chunks, but the controller executes only the first $H$ actions before observing the environment and replanning. This distinction allows replanning frequency to vary independently of the policy's prediction length.

\subsection{Simulator-Based Beam Search}
A search state contains the simulator configuration, including the agent coordinates, object position and orientation, and the dynamic variables needed to restore the state. At each planning point, we snapshot the current simulator state. Every hypothetical rollout starts from the appropriate restored state, and the root state is restored before any action is committed to the environment.

The task provides a coverage measure $c\in[0,1]$ describing overlap between the movable T and the target T. We use this feedback to rank candidate futures. Success is defined by at least 95\% coverage; an episode ends at success or after $T=300$ environment steps. Unlike real-world control, this setup provides both a simulator transition model and privileged access to task coverage.

At each expanded node, ACT predicts a nominal chunk. A candidate set contains this chunk and locally perturbed alternatives:
\begin{equation}
 A^{(j)}=\operatorname{clip}\!\left(A^{(0)}+\epsilon^{(j)}\right),
 \quad j=1,\ldots,C-1,
 \label{eq:noisy_candidates}
\end{equation}
where $C$ is the total number of candidate chunks, $\epsilon^{(j)}$ is action noise, and clipping enforces the workspace action bounds. Including the nominal chunk allows the controller to retain the policy prediction when the alternatives are worse.

Each candidate is simulated from its parent state, producing a child state scored by its coverage-based rollout outcome. At each level, beam search retains the $BW$ highest-scoring nodes and expands them until depth $D$, success, or the episode limit. The selected action is the root chunk leading to the best-scoring explored outcome. The controller restores the real state and commits its first $H$ actions. Only committed transitions contribute to episode reward; hypothetical search rollouts do not.

The execution horizon determines how often the controller replans. For $T=300$ and $H=10$, full beam search can be invoked at most
\begin{equation}
 N_{\mathrm{calls}}=\left\lceil T/H\right\rceil=30
 \label{eq:beam_calls}
\end{equation}
times per episode. Increasing $H$ lowers the number of planning calls but also delays opportunities to correct a poor action sequence.

\subsection{One Step Ahead}
We measure how often search changes the policy prediction through the \emph{alternative selection ratio} (ASR):
\begin{equation}
 \mathrm{ASR}=
 \frac{\text{number of selected perturbed chunks}}
      {\text{total number of chunk selections}}.
 \label{eq:asr}
\end{equation}
ASR is reported as a percentage. When search repeatedly selects the nominal chunk, evaluating all alternatives adds cost without changing the committed action.

OSA uses a preliminary rollout to decide whether full search is needed. Let $c_t$ be the current coverage and $\hat c_{\mathrm{nom}}$ the coverage at the end of the simulated nominal chunk. We define
\begin{equation}
 \Delta c=c_t-\hat c_{\mathrm{nom}}.
 \label{eq:coverage_drop}
\end{equation}
Full beam search is triggered only when
\begin{equation}
 \Delta c>\delta,\qquad \delta=0.05.
 \label{eq:one_step_threshold}
\end{equation}
Otherwise, the nominal chunk is selected directly. Here, ``one step'' refers to one candidate-chunk rollout, not a single low-level control step. The gate still incurs the cost of this rollout, but can avoid the remaining tree expansion. Algorithm~\ref{algo:one_step_ahead} describes the decision procedure.

\begin{algorithm}[t]
 \caption{One Step Ahead Action Selection}
 \label{algo:one_step_ahead}
 \begin{algorithmic}[1]
  \REQUIRE Observation $o_t$, simulator state $x_t$, frozen policy $\pi_\theta$, threshold $\delta=0.05$
  \ENSURE Selected action chunk $A_t^*$
  \STATE Save root simulator state $x_t$
  \STATE Predict $A_t^{(0)}\leftarrow\pi_\theta(o_t)$
  \STATE Measure current coverage $c_t$
  \STATE Simulate the nominal chunk from $x_t$
  \STATE Measure final simulated coverage $\hat c_{\mathrm{nom}}$
  \STATE Restore root state $x_t$
  \IF{$c_t-\hat c_{\mathrm{nom}}>\delta$}
    \STATE $A_t^*\leftarrow\operatorname{BeamSearch}(x_t,o_t,\pi_\theta)$
  \ELSE
    \STATE $A_t^*\leftarrow A_t^{(0)}$
  \ENDIF
  \STATE Restore $x_t$ and execute the first $H$ actions of $A_t^*$
 \end{algorithmic}
\end{algorithm}

The gate detects predicted regression rather than all forms of policy failure. In particular, a nominal chunk that makes no progress can bypass search, and a small final coverage drop can hide a harmful intermediate transition. OSA is therefore a heuristic for allocating computation, with no guarantee of optimality or safe execution.

\section{Experiments}
\subsection{Evaluation Setup}
We use the pretrained \nolinkurl{aadarshram/act_pusht} checkpoint with frozen parameters. Each configuration is evaluated on 50 PushT episodes with a maximum length of 300 steps. We compare direct ACT inference, full beam search, and OSA-gated beam search. The tree-structure ablation varies depth $D$, total candidate count $C$, and beam width $BW$. The replanning-frequency ablation varies the execution horizon $H$, and the OSA ablation uses a coverage-drop threshold of 0.05.

We report average cumulative reward, average maximum reward, task success, ASR, and runtime. For an episode of $\tau\leq300$ executed steps, cumulative reward is $\sum_{t=1}^{\tau}r_t$ and maximum reward is $\max_{1\leq t\leq\tau}r_t$. These episode metrics are averaged across the evaluation set. Cumulative reward is an accumulated score, not a coverage percentage. Success is the fraction of episodes that meet the 95\% coverage criterion. Runtime values describe the reported evaluation setup; no claim of real-time physical control is made.

\begin{table*}[t]
 \caption{PushT beam-search configurations, averaged over 50 episodes. $D$: tree depth; $C$: total candidate count per expanded node, including the nominal chunk; $BW$: beam width. Reward columns report episode-level cumulative and maximum reward, respectively. ASR is the percentage of selections using a perturbed chunk.}
 \label{tab:pusht-search-configurations}
 \centering
 \begin{tabular}{@{}lrrrccrrr@{}}
  \toprule
  Method & $D$ & $C$ & $BW$ & Avg. sum reward & Avg. max reward & Success (\%) & ASR (\%) & Runtime (s) \\
  \midrule
  Beam search & 2 & 4 & 2 & 90.9 & 0.77 & \textbf{18} & 56.4 & 14 \\
  Beam search & 4 & 2 & 2 & 90.8 & 0.78 & 16 & 37.5 & 10 \\
  Beam search & 4 & 4 & 2 & 90.9 & 0.77 & \textbf{18} & 58.23 & 10 \\
  Beam search & 4 & 8 & 2 & 94.7 & 0.77 & \textbf{18} & 75.6 & 33 \\
  Beam search & 4 & 8 & 4 & \textbf{103.2} & \textbf{0.81} & \textbf{18} & 75.0 & 47 \\
  Beam search & 4 & 8 & 8 & 92.2 & 0.80 & 10 & 78.5 & 87 \\
  Beam search & 8 & 4 & 2 & 90.9 & 0.77 & \textbf{18} & 58.23 & 16 \\
  \midrule
  ACT baseline & -- & -- & -- & 39.44 & 0.36 & 0 & -- & 2 \\
  \bottomrule
 \end{tabular}
\end{table*}

\begin{figure}[t]
 \centering
 \includegraphics[width=\columnwidth]{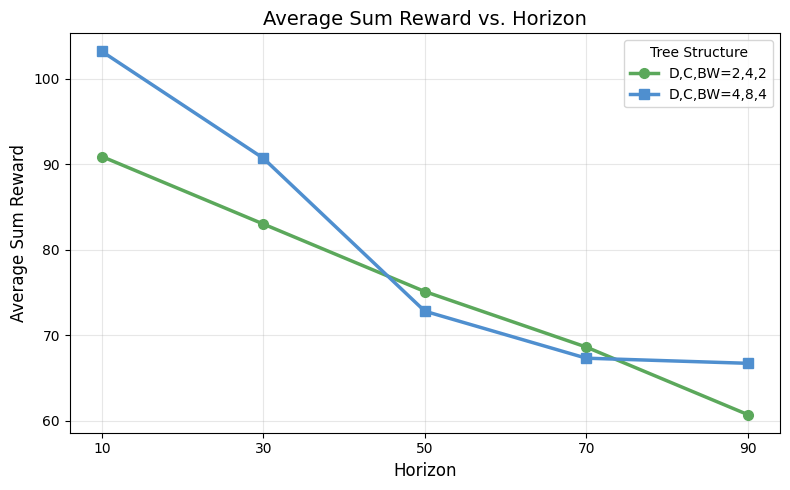}
 \caption{Average cumulative reward versus execution horizon $H$ for two beam-search configurations. Shorter horizons invoke search more frequently. Each point summarizes 50 episodes.}
 \label{fig:pusht-search-frequency}
\end{figure}

\begin{figure*}[t]
 \centering
 \includegraphics[width=\textwidth]{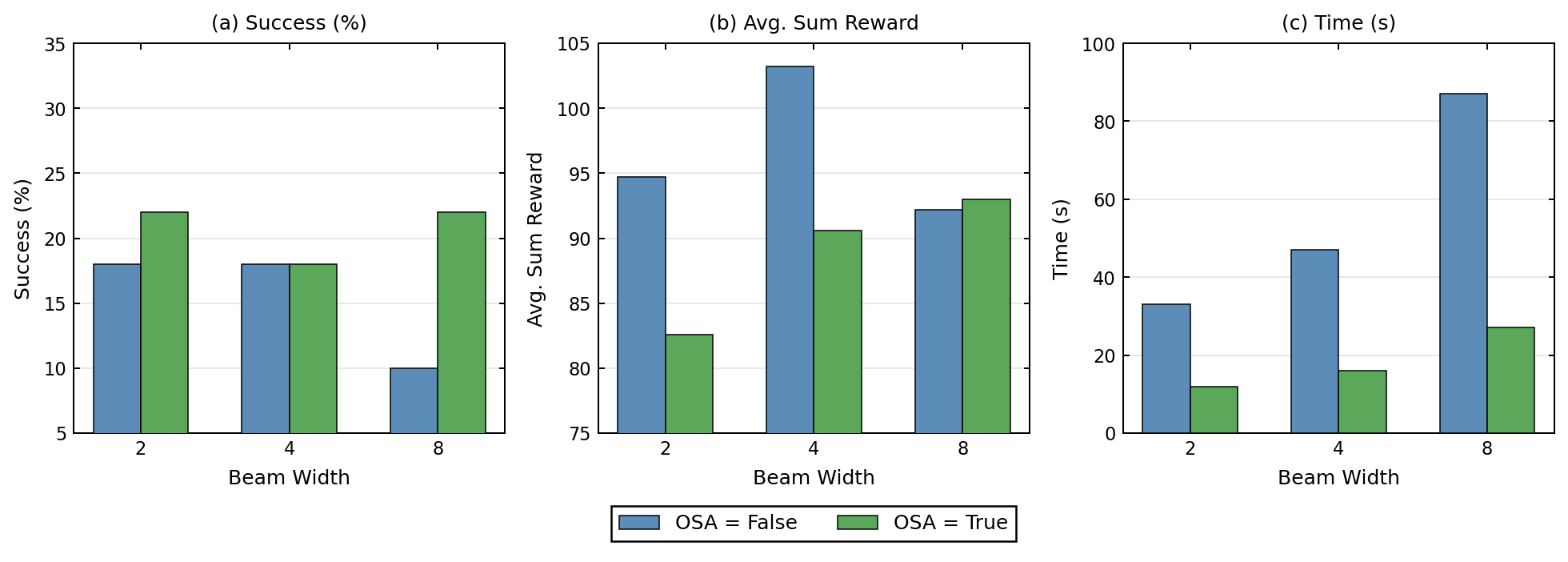}
 \caption{One Step Ahead ablation at $D=4$, $C=8$, and beam widths 2, 4, and 8. Blue: full beam search; green: OSA. Panels show success, average cumulative reward, and runtime. OSA preserves or improves success in these comparisons while reducing computation, although cumulative reward can decrease.}
 \label{fig:pusht-one-step-ahead}
\end{figure*}

\subsection{Effect of Tree Structure}
Table~\ref{tab:pusht-search-configurations} shows that beam search improves substantially over the evaluated ACT baseline. The configuration with $D=4$, $C=8$, and $BW=4$ achieves the highest average cumulative reward, 103.2 versus 39.44 for direct inference. Its average maximum reward rises from 0.36 to 0.81, and success improves from 0\% to 18\%. This comes at a runtime increase from 2 to 47 seconds, or approximately $23.5\times$.

Increasing depth alone yields little improvement. With $C=4$ and $BW=2$, depths 2, 4, and 8 produce the same reported average reward values and 18\% success. Increasing the number of candidates generally increases ASR: at $D=4$, $BW=2$, ASR rises from 37.5\% with $C=2$ to 75.6\% with $C=8$. This shows that a larger candidate set more often changes the policy's action, but the corresponding gains in task success remain modest.

Wider beams also do not guarantee better outcomes. At $D=4$ and $C=8$, increasing $BW$ from 4 to 8 raises runtime from 47 to 87 seconds while reducing average cumulative reward from 103.2 to 92.2 and success from 18\% to 10\%. Search quality depends on the proposal distribution and coverage heuristic as well as on the number of explored nodes.

\subsection{Replanning Frequency}

Figure~\ref{fig:pusht-search-frequency} compares $D=2$, $C=4$, $BW=2$ with $D=4$, $C=8$, $BW=4$. More frequent replanning generally improves cumulative reward. As the horizon increases, the controller commits to longer action prefixes and loses opportunities to correct deviations. The larger tree performs best at short horizons, but its advantage disappears at longer horizons. These results indicate that frequent feedback can matter more than deeper or wider search at infrequent decision points.

\subsection{Selective Search with One Step Ahead}
Figure~\ref{fig:pusht-one-step-ahead} compares OSA with full beam search. The preliminary coverage check reduces runtime by approximately $3\times$ in the reported ablations. Success is preserved or improved in the displayed settings, even when cumulative reward decreases. These metrics need not move together: cumulative reward measures progress throughout an episode, whereas binary success records whether the coverage threshold is reached.

The widest beam provides a useful example. Full search with $BW=8$ performs worse than the smaller beam, while OSA improves its success rate. One possible explanation is that skipping unnecessary expansion limits exposure to perturbed actions favored by a short-horizon heuristic. This is an interpretation of the observed behavior, rather than a separately measured causal effect.

OSA does not remove the cost of look-ahead: it still simulates the nominal chunk at every planning point and runs the full tree when the gate fires. Its benefit comes from avoiding redundant alternatives when the nominal prediction passes the coverage check.

\subsection{Qualitative Rollouts}

Figure~\ref{fig:pusht-rollout-comparison} illustrates how search changes the controller's decisions. Direct ACT inference fails to solve this episode within the step limit. Beam search and OSA instead select action chunks that bring the object into alignment. OSA bypasses full expansion at some decision points, including the initial state, while retaining search elsewhere. This example illustrates the gating mechanism; aggregate performance is reported in Table~\ref{tab:pusht-search-configurations} and Figure~\ref{fig:pusht-one-step-ahead}.

\section{Discussion and Limitations}
The experiments support selective search for a frozen action-chunking policy in PushT. They do not establish general improvements for VLA models: ACT is not language-conditioned, only one checkpoint and one task family are evaluated, and there is no direct comparison with VLA-Reasoner or VLAPS under a shared benchmark. Extending the controller to language-conditioned policies and tasks such as LIBERO is necessary to test that transfer.

The planner relies on restored simulator states and task-specific coverage feedback. Physical deployment would require a reliable dynamics model and a progress estimator available from real observations. Model errors could change both candidate rankings and the OSA trigger. The current evaluation does not quantify this sensitivity or validate the controller on real robots.

The nominal coverage gate is deliberately narrow. It can miss stalled behavior and does not detect every harmful action. The threshold is fixed, and the tree configuration is selected before a rollout. Adaptive budgets, uncertainty-aware triggers, and alternative search algorithms may improve the trade-off between task performance and computation.

The quantitative results summarize 50 episodes per configuration without confidence intervals or reported variability across independent seed sets. Small differences in success therefore warrant caution. In addition, the reported baseline has zero success, and the strongest full-search configuration reaches only 18\%; these gains do not imply that the task is solved reliably. Further evaluation with stronger checkpoints and matched compute budgets would clarify when search is preferable to improving the base policy.

\section{Conclusion}
We study look-ahead action selection motivated by search for VLA control, using a frozen ACT policy in PushT. Beam search over perturbed chunks improves success and cumulative reward over the evaluated baseline, but larger trees do not consistently improve performance. One Step Ahead reduces redundant tree expansion by checking the predicted coverage change of the nominal policy chunk before searching. In the reported ablations, this gate preserves or improves success while reducing runtime. These findings motivate selective planning for robotic policies and evaluation on VLAs and physical robots.

\section*{Acknowledgements}
This study was supported by the Scientific and Technological Research Council of T\"urkiye (TUBITAK), Grant No.\ 124E055.

\section*{Impact Statement}
This work studies inference-time action selection in a simulated manipulation task. Better allocation of planning computation may make robotic policies more efficient to evaluate and deploy. The experiments do not establish safety for physical robots: simulator access and privileged progress feedback do not account for all consequences of real-world actions. Deployment would require validation of the transition model, task feedback, and controller behavior in the intended operating environment.

\section*{Project Resources}
The project website is available at \url{https://azeer.co/research/lookahead/}. The PushT search implementation is available at \url{https://github.com/AhmedZeer/lerobot-tree/tree/main/examples/tree_search/pusht}.

The pretrained ACT checkpoint is available at \url{https://huggingface.co/aadarshram/act_pusht}. The experiments use the existing PushT benchmark; no new training dataset is introduced. The authors declare no competing interests.

\bibliographystyle{icml2026}
\bibliography{sn-bibliography}

\end{document}